\documentclass[12pt]{amsart}
\usepackage[utf8]{inputenc}
\usepackage[utf8]{inputenc}
\usepackage[T1]{fontenc}
\usepackage[all]{xy}
\usepackage{lipsum}
\usepackage{url}
\usepackage{tikz}
\usepackage{stackrel}
\usepackage{color}
\def\red{}

\usepackage{amsmath,amsthm,amssymb}

\usepackage{xcolor}

\usepackage{graphicx}

\newtheorem{theorem}{Theorem}[section]

\newtheorem{example}[theorem]{Example}
\newtheorem{formulation}[theorem]{Formulation}
\newtheorem{proposition}[theorem]{Proposition}
\theoremstyle{definition}
\newtheorem{definition}[theorem]{Definition}

\newtheorem{remark}[theorem]{Remark}

\numberwithin{equation}{section}

\begin{document}

\renewcommand{\bf}{\bfseries}
\renewcommand{\sc}{\scshape}

\title[Higher motion planning algorithms]%
{Multitasking collision-free optimal motion planning algorithms in Euclidean spaces}

\author{Cesar A. Ipanaque Zapata}
\address{Departamento de Matem\'{a}tica,Universidade de S\~{a}o Paulo, 
Instituto de Ci\^{e}ncias Matem\'{a}ticas e de Computa\c{c}\~{a}o -
USP , Avenida Trabalhador S\~{a}o-carlense, 400 - Centro CEP:
13566-590 - S\~{a}o Carlos - SP, Brasil}
\curraddr{Departamento de Matem\'{a}ticas, Centro de Investigaci\'{o}n y de Estudios Avanzados del I. P. N.
Av. Instituto Polit\'{e}cnico Nacional n\'{u}mero 2508,
San Pedro Zacatenco, Mexico City 07000, M\'{e}xico}
\email{cesarzapata@usp.br}
\thanks{The first author would like to thank grant\#2018/23678-6, S\~{a}o Paulo Research Foundation (FAPESP) for financial support.}

\author{Jes\'{u}s Gonz\'{a}lez}
\address{Departamento de Matem\'{a}ticas, Centro de Investigaci\'{o}n y de Estudios Avanzados del I. P. N.
Av. Instituto Polit\'{e}cnico Nacional n\'{u}mero 2508,
San Pedro Zacatenco, Mexico City 07000, M\'{e}xico}
\email{jesus@math.cinvestav.mx}



\begin{abstract} We present optimal motion planning algorithms which can be used in designing practical systems controlling objects moving in Euclidean space without collisions. Our algorithms are optimal in a very concrete sense, namely, they have the minimal possible number of local planners. Our algorithms are motivated by those presented by Mas-Ku and Torres-Giese (as streamlined by Farber), and are developed within the more general context of the multitasking (a.k.a.~higher) motion planning problem. In addition, an eventual implementation of our algorithms is expected to work more efficiently than previous ones when applied to systems with a large number of moving objects.
\end{abstract}

\maketitle


\section{Introduction}\label{secintro}

Let $X$ be the space of all possible \red{obstacle-free} configurations or states of a given autonomous system. For $n\geq 2,$  an \emph{$n$-th sequential motion planning algorithm on $X$} is a function which\red{,} to any $n$-tuple of configurations $(C_1,\ldots,C_n)\in X^n=X\times\cdots\times X$ ($n$ times)\red{,} assigns a continuous motion $\mu$ of the system, so that $\mu$ starts at the given initial state $C_1$, ends at the final desired state $C_n$, and passes sequentially through the additional $n-2$ prescribed intermediate states $C_2,\ldots,C_{n-1}$. The fundamental problem in robotics, \textit{the motion planning problem}, \red{deals with how to provide,} to any given autonomous system, with an $n$-th sequential motion planning algorithm. 

\medskip
For practical purposes, an $n$-th sequential motion planning algorithm should depend continuously on the $n$-tuple of points $(C_1,\ldots,C_n)$. Indeed, if the autonomous system performs within a noisy environment, absence of continuity could lead to instability issues in the behavior of the motion planning algorithm. Unfortunately, a (global) continuous $n$-th sequential motion planning algorithm on a space $X$ exists if and only if $X$ is contractible. Yet, if $X$ is not contractible, we could care about finding \emph{local} continuous $n$-th sequential motion planning algorithms, i.e., motion planning algorithms \red{$s$} defined only on a certain open set of $X^n$, \red{to which we refer as the domain of definition of $s$}. In these terms, a \emph{motion planner on $X$} is a set of local continuous $n$-th sequential motion planning algorithms whose \red{domains of definition} cover $X^n$. The \emph{$n$-th sequential topological complexity of $X$}, TC$_n(X)$, is then the minimal cardinality among motion planners on $X$, while a motion planner on $X$ is said to be \emph{optimal} if its cardinality is TC$_n(X)$. The design of explicit motion planners that are reasonably close to optimal is one of the challenges of modern robotics (see, for example Latombe \cite{latombe2012robot} and LaValle \cite{lavalle2006planning}). 

\medskip\red{In more detail, the components of the multitasking motion planning problem via topological complexity are as follows:}
\begin{formulation} \red{Ingredients in the multitasking motion planning problem via topological complexity:
\begin{enumerate}
    \item The obstacle-free configuration space $X$. The topology of this space is assumed to be fully understood in advance. 
    \item Query $n$-tuples $C=(C_1,C_2,\ldots,C_{n-1},C_n)\in X^n$. The point $C_1{}\in X$ is designated as the initial configuration of the query. The $n-2$ points  $C_2,\ldots,C_{n-1}$ in $X$ are designated as the prescribed intermediate configurations. The point $C_n{}\in X$ is designated as the goal configuration.  
\end{enumerate}
In the above setting, the goal is to either describe an $n$-th sequential motion planning algorithm, i.e., describe
\begin{enumerate}\addtocounter{enumi}{2}
\item An open covering $U=\{U_1,\ldots,U_k\}$ of $X^n$;
\item For each $i\in\{1,\ldots,k\}$, an $n$-th sequential planner, i.e., a continuous map $s_i\colon U_i\to P(X)$ satisfying $$s_i(C)\left(\dfrac{j}{n-1}\right)=C_{j+1},\quad 0\leq j\leq n-1$$ for any $C=(C_1,\ldots,C_n){}\in U_i$ (here $P(X)$ stands for the free-path space on $X$),
\end{enumerate}
or, else, report that such an algorithm does not exist.}
\end{formulation}

\medskip
Investigation of the problem of simultaneous collision-free sequential motion planners for $k$ distinguishable robots, each with state space $X$, leads us to study the \textit{ordered configuration space} $F(X,k)$ of $k$ distinct points on $X$ (see \cite{fadell1962configuration}). Explicitly, \[F(X,k)=\{(x_1,\ldots,x_k)\in X^k\mid ~~x_i\neq x_j\text{ for } i\neq j \},\] topologised as a subspace of the Cartesian power $X^k$. \red{For our purposes, we ignore dynamics and other differential constraints, and we  focus primarily on the translations required to move the robot. So we will have in mind an infinitesimal mass particle as an object (e.g., infinitesimally small ball). Namely, we consider our robots as points in the Euclidean space $\mathbb{R}^d$. The configuration space of each robot is determined by its position in $\mathbb{R}^d$. In other words, $X=\mathbb{R}^d$.} Note that the $i$-th coordinate of a point $(x_1,\ldots,x_n)\in F(\mathbb{R}^d,k)$ represents the state or position of the $i$-th moving object, so that the condition $x_i\neq x_j$ reflects the collision-free requirement. Thus, a (local) $n$-th sequential motion planning algorithm in $F(\mathbb{R}^d,k)$ assigns to any $n$-tuple of configurations $(C_1,\ldots,C_n)$ in (an open set of) $F(\mathbb{R}^d,k)\times\cdots\times F(\mathbb{R}^d,k)$ a continuous curve of configurations \[\Gamma(t)\in F(\mathbb{R}^d,k),~~t\in [0,1],\]  such that $\Gamma\left(\dfrac{i}{n-1}\right)=C_{i+1}$ for $0\leq i\leq n-1$. 

\medskip
In this work we present two $n$-th sequential motion planners in $F(\mathbb{R}^d,k)$ for any $n\geq 2$. \red{Inspired by the work done for $n=2$ by Farber (\cite{farber2017configuration}) and Mas-Ku and Torres-Giese (\cite{hugo2015}), we present two $n$-th sequential motion planners in $F(\mathbb{R}^d,k)$ for any $n\geq 2$.} The first planner has $n(k-1)+1$ \red{domains of definition,} works for any $d,n,k\geq 2$, and is optimal if $d\geq 3$ is odd (in view of Theorem~\ref{gongra} below). The second planner, which is defined only for $d\geq2$ even, has $n(k-1)$ regions of continuity and is optimal too (again by Theorem~\ref{gongra}). The motion planning algorithms we present in this work are easily implementable in practice, and (for $n=2$) work more efficiently than those of Farber when the number $k$ of moving objects becomes large (see Remark~\ref{muchasparticulas}). 

\medskip
\red{Despite the multitasking motion planning problem is relatively new, its theoretical properties via topological complexity (a la Farber) have been studied intensively. Yet, concrete algorithms are scarce (only those coming from~\cite{farber2017configuration} and~\cite{hugo2015}), while specific implementations are inexistent. In fact, this work takes a first step in the direction of producing explicit algorithms.}

\section{Preliminary results}

The concept of $n$-th sequential topological complexity (also called $n$-th ``higher'' TC) was introduced by Rudyak in \cite{rudyak2010higher}, and further developed in~\cite{BGRT}. Here we recall the basic definitions and properties.

\medskip
For a topological space $X$, let $P(X)$ denote the space of free paths on $X$ with the compact-open topology. For $n\geq 2$, consider the evaluation fibration \begin{equation}\label{evaluation-fibration}
    e_n:P(X)\to X^n,~e_n(\gamma)=\left(\gamma(0),\gamma\left(\dfrac{1}{n-1}\right),\ldots,\gamma\left(\dfrac{n-2}{n-1}\right),\gamma(1)\right).
\end{equation} 
An \textit{$n$-th sequential motion  planning  algorithm} is  a  section $s\colon X^n\to P(X)$ of  the  fibration  $e_n$, i.e.,~ a (not necessarily continuous) map satisfying $e_n\circ s=id_{X^n}$. A continuous $n$-th sequential motion planning algorithm in $X$ exists if and only if the space $X$ is contractible, which forces the following definition. The \textit{$n$-th sequential topological complexity} TC$_n(X)$ of a path-connected space $X$ is the Schwarz genus of the evaluation fibration~(\ref{evaluation-fibration}). In  other  words the $n$-th sequential topological complexity of $X$ is the smallest positive integer TC$_n(X)=k$ for which  the product $X^n$ is covered by $k$ open subsets $X^n=U_1\cup\cdots\cup U_k$ such that for any $i=1,2,\ldots,k$ there exists a continuous section $s_i:U_i\to P(X)$ of $e_n$ 
over $U_i$ (i.e., $e_n\circ s_i=id$).

\red{
\begin{example}
Suppose that $X$ is a convex subset of a Euclidean space $\mathbb{R}^d$. Given an $n$-tuple
of configurations $(C_1,\ldots,C_n)\in X^n$, we may move with constant velocity along the
straight line segment connecting $C_i$ and $C_{i+1}$ for each $i=1,\ldots,n-1$. This clearly produces a continuous algorithm
for the $n$-th sequential motion planning problem in $X$. Thus we have TC$_n(X)=1$.
\end{example}}

\medskip
Note that TC$_2$ coincides with Farber`s topological complexity, which is defined in terms of motion planning algorithms for a robot moving between initial-final configurations~\cite{farber2003topological}. The more general TC$_n$ is Rudyak`s higher topological complexity of motion planning problem, whose input requires, in addition of initial-final states, $n-2$ intermediate states of the robot. We will use the expression ``motion planning algorithm'' as a substitute of ``$n$-th sequential motion planning algorithm for $n=2$''. 

\medskip
\red{The} definition of TC$_n(X)$ deals with open subsets of $X^n$ admitting continuous sections of the evaluation fibration (\ref{evaluation-fibration}), \red{yet} for practical purposes, the construction of explicit $n$-th sequential motion planning algorithms is usually done by partitioning the whole space $X^n$ into pieces, over each of which \red{a continuous section for~(\ref{evaluation-fibration}) is set}. Since any such partition necessarily contains subsets which are not open (recall $X$ has been assumed to be path-connetected), we need to be able to operate with subsets of $X^n$ of a more general nature.

\begin{definition}
A topological space $X$ is a \textit{Euclidean Neighbourhood Retract} (ENR) if it can be embedded into an Euclidean space $\mathbb{R}^d$ with an open neighbourhood $U$, $X\subset U\subset \mathbb{R}^d$, admiting a retraction $r:U\to X,$ $r\mid_U=id_X$.
\end{definition}

\begin{example}
A subspace $X\subset \mathbb{R}^d$ is an ENR if and only if it is locally compact and locally contractible, see~\cite[Chap.~4, Sect.~8]{dold2012lectures}. This implies that all finite-dimensional polyhedra, manifolds and semi-algebraic sets are ENRs.
\end{example}

\begin{definition}
Let $X$ be an ENR. An $n$-th sequential motion planning algorithm $s:X^n\to P(X)$ is said to be \textit{tame} if $X^n$ splits as a pairwise disjoint union $X^n=F_1\cup\cdots\cup F_k$, where each $F_i$ is an ENR, and each restriction $s\mid_{F_i}:F_i\to P(X)$ is continuous. The subsets $F_i$ in such a decomposition are called \emph{domains of continuity} for $s$.
\end{definition}

\begin{proposition}\emph{(\cite[Proposition 2.2]{rudyak2010higher})}\label{rudi}
For an ENR $X$, TC$_n(X)$ is the minimal number of domains of continuity $F_1,\ldots,F_k$ for tame $n$-th sequential motion planning algorithms $s:X^n\to P(X)$.
\end{proposition}

\medskip
\red{Recall, that an $n$-th sequential motion  planning  algorithm $s=\{s_i:F_i\to P(X)\}_{i=1}^{\ell}$ is called optimal when $\ell=$TC$_n(X)$.}

\medskip
\red{Given an $n$-th sequential motion  planning  algorithm $s=\{s_i:F_i\to P(X)\}_{i=1}^{\ell}$ as above, one may organize the implementation as follows. Given an $n$-tuple of configurations $(C_1,\ldots,C_n)$, we first find the subset $F_i$ such that $(C_1,\ldots,C_n){}\in F_i$ and then we give the path $s_i(C_1,\ldots,C_n)$ as an output.}

\begin{remark}
In the final paragraph of the introduction we noted that in this paper we construct optimal $n$-th sequential motion planners in $F(\mathbb{R}^d,k)$. We can now be more precise: we actually construct $n$-th sequential tame motion planning algorithms with the advertized optimality property.
\end{remark}

Since~(\ref{evaluation-fibration}) is a fibration, the existence of a continuous motion planning algorithm on a subset $A$ of $X^n$ implies the existence of a corresponding continuous motion planning algorithm on any subset $B$ of $X^n$ deforming to $A$ within $X^n$. Such a fact is argued next in a constructive way, generalizing~\cite[Example 6.4]{farber2017configuration} (the latter given for $n=2$). This of course suits best our implementation-oriented objectives.

\begin{remark}[Constructing motion planning algorithms via deformations: higher case]\label{constructing-sections-via-deformations-higher-case}
 Let $s_A:A\to P(X)$ be a continuous motion planning algorithm defined on a subset $A$ of $X^n$. Suppose a subset $B\subseteq X^n$ can be continuously deformed within $X^n$ into $A$. Choose a homotopy $H:B\times [0,1]\to X^n$ such that $H(b,0)=b$ and $H(b,1)\in A$ for any $b\in B$. Let $h_1,\ldots, h_n$ be the Cartesian components of $H$, $H=(h_1,\ldots,h_n)$. As schematized in the picture
$$
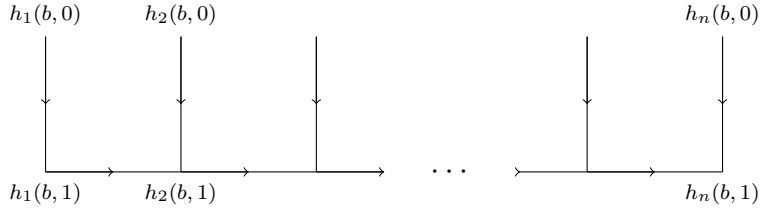
\begin{tikzpicture}[x=.6cm,y=.6cm]
\draw(0,3)--(0,0); \draw(3,3)--(3,0); \draw(6,3)--(6,0);\draw(12,3)--(12,0);
\draw(15,3)--(15,0); \draw(0,0)--(7.5,0); \draw(10.5,0)--(15,0); 
\draw[->](0,3)--(0,1.5); \draw[->](3,3)--(3,1.5); \draw[->](6,3)--(6,1.5);
\draw[->](10.5,0)--(10.51,0); \draw[->](12,0)--(13.5,0); \draw[->](12,3)--(12,1.5); \draw[->](15,3)--(15,1.5); 
\draw[->](0,0)--(1.5,0); \draw[->](3,0)--(4.5,0); \draw[->](6,0)--(7.5,0);  \node at (9,0) {$\cdots$};
\node [below] at (0,0) {\tiny$h_1(b,1)$}; \node [above] at (0,3) {\tiny$h_1(b,0)$};
\node [below] at (3,0) {\tiny$h_2(b,1)$}; \node [above] at (3,3) {\tiny$h_2(b,0)$};
\node [below] at (15,0) {\tiny$h_n(b,1)$}; \node [above] at (15,3) {\tiny$h_n(b,0)$};
\end{tikzpicture}
$$
(where $H$ runs from top to bottom and $s_A$ runs from left to right), the path $s_A(H(b, 1))$ connects in sequence the points $h_i(b,1)$, $1\leq i\leq n$, i.e., 
$$
s_A(H(b, 1))\left(\dfrac{i}{n-1}\right)=h_{i+1}(b,1), \quad 0\leq i\leq n-1,
$$
whereas the formula
$$
 s_B(b)(\tau) = \begin{cases}
    h_1(b,3(n-1)\tau), & \hbox{$0\leq \tau\leq \frac{1}{3(n-1)}$;} \\
    s_A(H(b,1))(3\tau-\frac{1}{n-1}), & \hbox{$\frac{1}{3(n-1)}\leq \tau\leq \frac{2}{3(n-1)}$;}\\
    h_2(b,3-3(n-1)\tau), & \hbox{$\frac{2}{3(n-1)}\leq \tau\leq \frac{1}{n-1}$;}\\
    h_2(b,3(n-1)\tau-3), & \hbox{$\frac{1}{n-1}\leq \tau\leq \frac{4}{3(n-1)}$;} \\
    s_A(H(b,1))(3\tau-\frac{3}{n-1}), & \hbox{$\frac{4}{3(n-1)}\leq \tau\leq \frac{5}{3(n-1)}$;}\\
    h_3(b,6-3(n-1)\tau), & \hbox{$\frac{5}{3(n-1)}\leq \tau\leq \frac{2}{n-1}$;}\\  \quad\vdots\\
    h_{n-1}(b,3(n-1)\tau-3(n-2)), & \hbox{$\frac{n-2}{n-1}\leq \tau\leq \frac{3n-5}{3(n-1)}$;} \\
    s_A(H(b,1))(3\tau-\frac{2n-3}{n-1}), & \hbox{$\frac{3n-5}{3(n-1)}\leq \tau\leq \frac{3n-4}{3(n-1)}$;}\\
    h_n(b,3(n-1)-3(n-1)\tau), & \hbox{$\frac{3n-4}{3(n-1)}\leq \tau\leq 1$,}
\end{cases}.
$$
 defines a continuous section $s_B:B\to P(X)$ of~(\ref{evaluation-fibration}) over $B$. Note that \begin{eqnarray}
 s_B(b) &=& h_1(b,-)\cdot s_A(H(b,1))\mid_1\cdot\ h_2(b,-)^{-1}\cdot\\ 
 & & h_2(b,-)\cdot  s_A(H(b,1))\mid_2\cdot\ h_3(b,-)^{-1}\cdot\cdots \cdot\ \nonumber\\ 
 & & h_{n-1}(b,-)\cdot s_A(H(b,1))\mid_{n-1}\cdot\ h_n(b,-)^{-1},\nonumber
 \end{eqnarray} where $s_A(H(b,1))\mid_j$ is the restriction of $s_A(H(b,1))$ to the segment $$\left[\dfrac{j-1}{n-1},\dfrac{j}{n-1}\right],$$ i.e., \[s_A(H(b,1))\mid_j(t)=s_A(H(b,1))\left(\dfrac{1}{n-1}\left(t-\dfrac{j-1}{n-1}\right)\right), \quad t\in [0,1],\]  for $j=1,\ldots,n-1$.
Summarizing: a deformation of $B$ into $A$ and a continuous motion planning algorithm defined on $A$ determine an explicit continuous motion planning algorithm defined on $B$.
\end{remark}

The final ingredient we need is the value of TC$_n(F(\mathbb{R}^d,k))$, computed by Gonz{\'a}lez and Grant in \cite{gonzalez2015sequential}.

\begin{theorem}\emph{(\cite{gonzalez2015sequential})}\label{gongra}For $d,k,n\geq2$,
\[\text{TC}_n(F(\mathbb{R}^d,k))=\left\{
  \begin{array}{ll}
     n(k-1)+1, & \hbox{if $d$ is odd;}\\
     n(k-1), & \hbox{if $d$ is even.}
    \end{array}
\right.\]
\end{theorem}



\section{Optimal tame motion planning algorithm in $F(\mathbb{R}^n,k)$}

In this section we make minor modifications in the tame motion planning algorithms described by Farber in~\cite{farber2017configuration} for $F(\mathbb{R}^d,k)$. As noted in the introduction, the first advantage of our streamlined algorithm is that an implementation will run more efficiently when the number $k$ of moving objects becomes large (see Remark~\ref{muchasparticulas}). The second advantage is that the streamlined algorithm generalizes to the multitasking (sequential) motion planning realm (Section~\ref{seccion4seqalgorithm}).

\subsection{Giese-Mas' motion planning algorithm in $F(\mathbb{R}^d,k)$ revisited}\label{section1}

\subsubsection{Section over $F(\mathbb{R},k)\times F(\mathbb{R},k)$}

\begin{figure}[h]
 \centering
\begin{tikzpicture}[x=.6cm,y=.6cm]
\draw[->](-2,0)--(-1,0); \draw[->](6,0)--(6,1); \draw[->](1,0)--(1,2); \draw[->](11,0)--(11,3);
\draw[->](-2,0)--(-2,1); \draw[->](6,1)--(14,1); \draw[->](1,2)--(9,2); \draw[->](11,3)--(4,3);
\draw(0,0)--(15,0); \draw[->](14,1)--(14,0); \draw[->](9,2)--(9,0); \draw[->](4,3)--(4,0);
\node [below] at (-1,0) {\tiny$e_1$};
\node [above] at (-2,1) {\tiny$e_2$};
\filldraw[color=black!60, fill=black!5, very thick](6,0) circle (0.5); \node[ ] at (6,0) {\tiny$1$}; \filldraw[color=black!60, fill=black!50, very thick](14,0) circle (0.5); \node[ ] at (14,0) {\tiny$1$};
\filldraw[color=black!60, fill=black!5, very thick](1,0) circle (0.5); \node[ ] at (1,0) {\tiny$2$}; \filldraw[color=black!60, fill=black!50, very thick](9,0) circle (0.5); \node[ ] at (9,0) {\tiny$2$};
\filldraw[color=black!60, fill=black!5, very thick](11,0) circle (0.5); \node[ ] at (11,0) {\tiny$3$}; \filldraw[color=black!60, fill=black!50, very thick](4,0) circle (0.5); \node[ ] at (4,0) {\tiny$3$};
\end{tikzpicture}
\caption{Section over $F(\mathbb{R},k)\times F(\mathbb{R},k)$. Vertical arrows pointing upwards (downwards) describe the first (last) third of the path $\Gamma^{C,C'},$ whereas horizontal arrows describe the middle third of $\Gamma^{C,C'}$.}
 \label{algorithm1}
\end{figure}
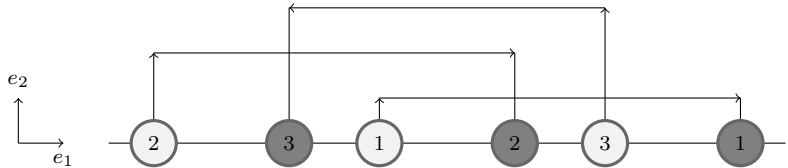

We think of $F(\mathbb{R},k)$ as a subspace of $F(\mathbb{R}^d,k)$ via the embedding $\mathbb{R}\hookrightarrow \mathbb{R}^d$, $x\mapsto (x,0,\ldots,0)$. Consider the first two standard basis elements $e_1=(1,0,\ldots,0)$ and $e_2=(0,1,0,\ldots,0)$ in $\mathbb{R}^d$ (we assume $d\geq2$). Given two configurations $C=(x_1,\ldots,x_k)$ and $C^\prime=(x^\prime_1,\ldots,x^\prime_k)$ in $F(\mathbb{R},k)$, let $\Gamma^{C,C'}$ be the path in $F(\mathbb{R}^d,k)$ from $C$ to $C^\prime$ depicted in Figure~\ref{algorithm1}.

\medskip

Explicitly, $\Gamma^{C,C'}$ has components $(\Gamma^{C,C'}_1,\ldots,\Gamma^{C,C'}_k)$ defined by 
\begin{equation}
    \label{Gamma1}\Gamma^{C,C'}_{i}(t)=\begin{cases}
    x_{i}+\red{(3ti)} e_2, & \hbox{for $0\leq t\leq \frac{1}{3}$;} \\
    x_{i}+ie_2+(3t-1)(x^\prime_{i}-x_{i}), & \hbox{for $\frac{1}{3}\leq t\leq \frac{2}{3}$;}\\
    x^\prime_{i}+i(3-3t)e_2, & \hbox{for $\frac{2}{3}\leq t\leq 1$}.
\end{cases}
\end{equation}
This yields a continuous motion planning algorithm $\Gamma:F(\mathbb{R},k)\times F(\mathbb{R},k)\to P F(\mathbb{R}^d,k)$. 

\begin{remark}\label{muchasparticulas}
The algorithm $\Gamma$ plays the role of the section $\sigma$ in~\cite[Equation~(18)]{farber2017configuration}. In that work, $\sigma$ is constructed via a concatenation process which, in our notation, involves having constructed, in advance, $(k!)^2$ paths. An implementation of this motion planning algorithm is bound to have complexity issues for large values of $k$ (i.e.,~when the number of moving particles is large). We avoid the problem with the explicit formula~(\ref{Gamma1}).
\end{remark}

\subsubsection{The sets $A_i$.}\label{subseccion312}

Let $p:\mathbb{R}^d\to \mathbb{R}, (y_1,\ldots,y_k)\mapsto(y_1,0,\ldots,0)$ denote the projection onto the first coordinate. For a configuration $C=(x_1,\ldots,x_k)\in F(\mathbb{R}^d,k)$, $\text{cp}(C)$ denotes the cardinality of the set of projection points $P(C) =\{p(x_1),\ldots,p(x_k)\}$. Note that $\text{cp}(C)$ ranges in $\{1, 2,\ldots,k\}$. Let $A_i$ denote the set of all configurations $C\in F(\mathbb{R}^d,k)$ with $\text{cp}(C)=i$. $A_i$ is an ENR, because it is a semi-algebraic set. Note that the closure of each set $A_i$ is contained in the union of the sets $A_j$ with $j\leq i$:
\begin{equation}\label{cerraduras1}
\overline{A_i}\subset \bigcup_{j\leq i}A_j.
\end{equation}

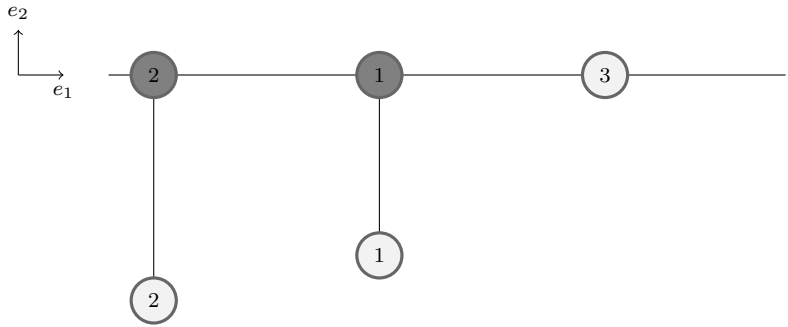
\begin{figure}[h]
 \centering
\begin{tikzpicture}[x=.6cm,y=.6cm]
\draw[->](-2,0)--(-1,0); 
\draw[->](-2,0)--(-2,1); 
\draw(0,0)--(15,0); 
\node [below] at (-1,0) {\tiny$e_1$};
\node [above] at (-2,1) {\tiny$e_2$};
 \draw[->](6,-4)--(6,0); \filldraw[color=black!60, fill=black!5, very thick](6,-4) circle (0.5); \node[ ] at (6,-4) {\tiny$1$}; \filldraw[color=black!60, fill=black!50, very thick](6,0) circle (0.5); \node[ ] at (6,0) {\tiny$1$};
 \draw[->](1,-5)--(1,0); \filldraw[color=black!60, fill=black!5, very thick](1,-5) circle (0.5); \node[ ] at (1,-5) {\tiny$2$}; \filldraw[color=black!60, fill=black!50, very thick](1,0) circle (0.5); \node[ ] at (1,0) {\tiny$2$};
 \filldraw[color=black!60, fill=black!5, very thick](11,0) circle (0.5);\node[ ] at (11,0) {\tiny$3$};
\end{tikzpicture}
\caption{Linear deformation of $A_k$ onto $F(\mathbb{R},k)$ inside $F(\mathbb{R}^d,k)$.}
 \label{algorithm2}
\end{figure}

\begin{remark}\label{thesetAk}
The map $\varphi=(\varphi_1,\ldots,\varphi_k):A_k\times [0,1]\to F(\mathbb{R}^d,k)$ given by the formula 
\begin{equation}
    \label{linear-transformation} \varphi_i(C,t)=x_i+t(p(x_i)-x_i),~i=1,\ldots,k,
\end{equation}
where $C=(x_1,\ldots,x_k)\in A_k$, defines a continuous deformation of $A_k$ onto $F(\mathbb{R},k)$ inside $F(\mathbb{R}^d,k)$ (see Figure~\ref{algorithm2}). In particular, $\Gamma$ and the $n=2$ case in Remark~\ref{constructing-sections-via-deformations-higher-case} yield a continuous motion planning algorithm defined on $A_k\times A_k$.
\end{remark}

\medskip

\begin{figure}[h]
 \centering
\begin{tikzpicture}[x=.6cm,y=.6cm]
\draw[->](-2,0)--(-1,0); 
\draw[->](-2,0)--(-2,1); 
\draw(0,0)--(15,0); 
\node [below] at (-1,0) {\tiny$e_1$};
\node [above] at (-2,1) {\tiny$e_2$};
 \filldraw[color=black!60, fill=black!50, very thick](11,0) circle (0.5);\node[ ] at (11,0) {\tiny$1$};
 \draw[->](4,-5)--(6,-5); \filldraw[color=black!60, fill=black!5, very thick](4,-5) circle (0.5); \node[ ] at (4,-5) {\tiny$2$};\filldraw[color=black!60, fill=black!50, very thick](6,-5) circle (0.5); \node[ ] at (6,-5) {\tiny$2$};
 \draw[->](4,5)--(8,5); \filldraw[color=black!60, fill=black!5, very thick](4,5) circle (0.5); \node[ ] at (4,5) {\tiny$3$};
 \filldraw[color=black!60, fill=black!50, very thick](8,5) circle (0.5); \node[ ] at (8,5) {\tiny$3$};
 \draw[dashed](4,-6)--(4,6);
\end{tikzpicture}
\caption{Desingularization.}
 \label{algorithm3}
\end{figure}
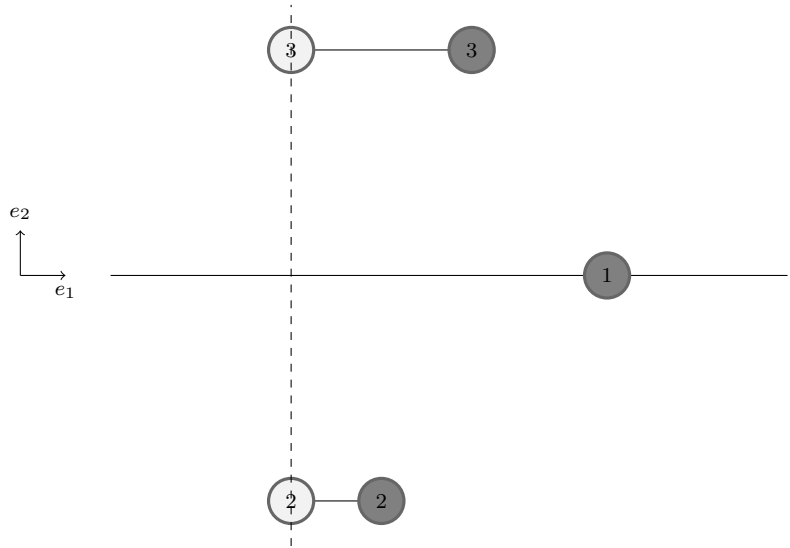

For a configuration $C=(x_1,\ldots,x_k)\in A_i$, set
\[\epsilon (C):=\begin{cases}\frac{1}{k}\min\{\mid p(x_r)-p(x_s)\mid:~ p(x_r)\neq p(x_s)\},& \mbox{if $i\geq2$;} \\
 1, & \mbox{if $i=1$.} \end{cases}\]
In addition, for $C$ as above and $t\in[0,1]$, set
\[D^i(C,t)=\begin{cases}(z_1(C,t),\ldots,z_k(C,t)), & \mbox{if $i<k$;}\\
 C, & \mbox{if $i=k$,}
 \end{cases}\]
 where $z_j(C, t)=x_j+t(j-1)\epsilon(C)e_1$ for $j=1,\ldots,k$. This defines a continuous ``desingularization'' deformation $D^i:A_i\times [0,1]\to F(\mathbb{R}^d,k)$ of $A_i$ into $A_k$ inside $F(\mathbb{R}^d,k)$ (see Figure~\ref{algorithm3}). As in Remark~\ref{thesetAk}, this yields a continuous motion planning algorithm on any subset $A_i\times A_j$, for $i,j\in\{1,\ldots,k\}$.
 
 \medskip

\subsubsection{Combining regions of continuity.} 
\begin{figure}[h]
 \centering
\begin{tikzpicture}[x=.6cm,y=.6cm]
\draw[->](-2,0)--(-1,0); 
\draw[->](-2,0)--(-2,1); 
\draw(0,0)--(17,0); 
\node [below] at (-1,0) {\tiny$e_1$};
\node [above] at (-2,1) {\tiny$e_2$};
 \draw[->](11,0)--(11,1); \draw[->](11,1)--(14,1); \draw[->](14,1)--(14,0); \draw[->](14,0)--(14,5); 
 \filldraw[color=black!60, fill=black!5, very thick](11,0) circle (0.5);\node[ ] at (11,0) {\tiny$1$};
 \filldraw[color=black!60, fill=black!50, very thick](14,5) circle (0.5);\node[ ] at (14,5) {\tiny$1$};
 \draw[->](4,-5)--(6,-5);\draw[->](6,-5)--(6,0); \draw[->](6,0)--(6,2); \draw[->](6,2)--(15,2); \draw[->](15,2)--(15,0); \draw[->](15,0)--(14,0);  
 \filldraw[color=black!60, fill=black!5, very thick](4,-5) circle (0.5); \node[ ] at (4,-5) {\tiny$2$};\filldraw[color=black!60, fill=black!50, very thick](14,0) circle (0.5); \node[ ] at (14,0) {\tiny$2$};
 \draw[->](4,5)--(8,5);\draw[->](8,5)--(8,0); \draw[->](8,0)--(8,3); \draw[->](8,3)--(16,3); \draw[->](16,3)--(16,0); \draw[->](16,0)--(16,-5); \draw[->](16,-5)--(14,-5);  
 \filldraw[color=black!60, fill=black!5, very thick](4,5) circle (0.5); \node[ ] at (4,5) {\tiny$3$};
 \filldraw[color=black!60, fill=black!50, very thick](14,-5) circle (0.5); \node[ ] at (14,-5) {\tiny$3$};
 \draw[dashed](4,-4.5)--(4,4.5);
 \draw[dashed](14,-0.5)--(14,-4.5);
\end{tikzpicture}
\caption{The motion planning algorithm in $F(\mathbb{R}^d,k)$.}
 \label{algorithm4}
\end{figure}
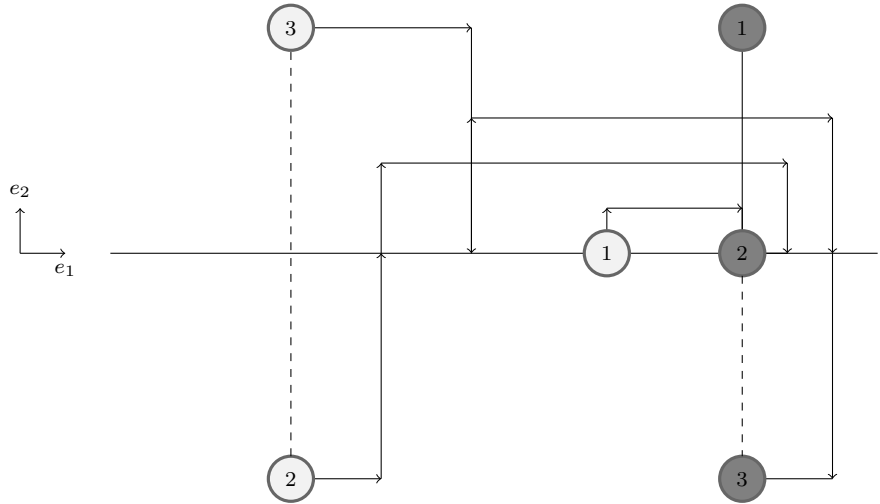

We have constructed continuous motion planning algorithms
\[\sigma_{i,j}\colon A_i\times A_j\to PF(\mathbb{R}^d,k),\quad i,j=1, 2,\ldots,k,\] by applying iteratively the construction in Remark~\ref{constructing-sections-via-deformations-higher-case}. For $i,j\in\{1, 2,\ldots,k\}$, the  sets $A_i\times A_j$ are  pairwise disjoint ENR's covering $F(\mathbb{R}^d,k)\times F(\mathbb{R}^d,k)$. The resulting estimate TC$(F(\mathbb{R}^d,k))\leq k^2$ is next improved by noticing that the sets $A_i\times A_j$  can be repacked into $2k-1$ pairwise disjoint ENR's each admitting its own continuous motion planning algorithm. Indeed,~(\ref{cerraduras1}) implies that $A_i\times A_j$ and $A_{i'}\times A_{j'}$ are ``topologically disjoint'' in the sense that $\overline{A_i\times A_j}\cap (A_{i^\prime}\times A_{j^\prime})=\varnothing$, provided $i+j=i'+j'$ and $(i,j)\neq(i',j')$. Consequently, for $2\leq\ell\leq 2k,$ the motion planning algorithms $\sigma_{i,j}$ having $i+j=\ell$ determine a (well-defined) continuous motion planning algorithm on the ENR
\begin{equation*}
    W_{\ell}=\bigcup_{i+j=\ell}A_i\times A_j.
\end{equation*}
We have thus constructed a (global) tame motion planning algorithm in $F(\mathbb{R}^d,k)$ having the $2k-1$ domains of continuity $W_2,W_3,\ldots, W_{2k}$ (see Figure~\ref{algorithm4}).

\medskip


\subsection{Farber's motion planning algorithm in $F(\mathbb{R}^{\red{2d}},k)$ revisited}\label{section2}

We now improve the motion planning algorithm in $F(\mathbb{R}^d,k)$ of the previous section under the assumption (in force throughout this subsection) that $d\geq 2$ is even. The improved motion planning algorithm will have $2k-2$ domains of continuity.

\smallskip
The first steps are nearly identical to those in the previous subsection: For a configuration $C=(x_1,\ldots,x_k)\in F(\mathbb{R}^d,k)$, consider the affine line $L_C$ through the points $x_1$ and $x_2$, oriented in the direction of the unit vector \[e_C=\dfrac{x_2-x_1}{\mid x_2-x_1\mid},\] and let $L^\prime_C$ denote the line passing through the origin and parallel to $L_C$ (with the same orientation as $L_C$). Let $p_C:\mathbb{R}^d\to L_C$ be the orthogonal projection, and let $\overline{\text{cp}}(C)$ be the cardinality of the set $\{p_C(x_1),\ldots,p_C(x_k)\}$. Note that $\overline{\text{cp}}(C)$ ranges in $\{2,\ldots, k\}$. For $i\in\{2,\ldots,k\}$, let $A_i$ denote\footnote{Beware that $A_i$ stands for a different set than the set with the same notation in Subsection~\ref{section1}.} the set of all configurations $C\in F(\mathbb{R}^d,k)$ with $\overline{\text{cp}}(C)=i$. The various $A_i$ are ENR's satisfying
\begin{equation}\label{cerraduras2}
\overline{A_i}\subset \bigcup_{j\leq i}A_j.
\end{equation}

\begin{figure}[htb]
 \centering
 \begin{tikzpicture}[x=.6cm,y=.6cm]
\draw[->](0,0)--(1,0.5); \node[below] at (0,0) {\tiny$0$}; \node[above] at (1,0.5) {\tiny$e_C$};
\draw (-6,0)--(8,7); \node[below] at (-5,0.5) {\tiny$x_1$}; \node[below] at (-2,2) {\tiny$x_2$};
\node[above] at (8,7) {\tiny$L_C$}; \draw[->](4,7)--(5,5.5);
\node[above] at (4,7) {\tiny$x_i$}; \node[below] at (5,5.5) {\tiny$p_C(x_i)$}; 
\end{tikzpicture}
\caption{The line $L_C$, its orientation $e_C$, and the projection $p_C$.}
\label{Lc}
\end{figure}
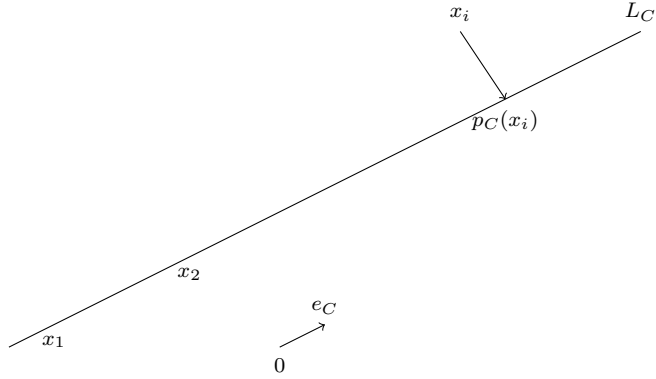

\subsubsection{Desingularization}
For a configuration $C=(x_1,\ldots,x_k) \in A_i$, set
\[\overline{\epsilon} (C):=\dfrac{1}{k}\min\{\mid p_{C}(x_r)-p_{C}(x_s)\mid\colon  p_{C}(x_r)\neq p_{C}(x_s)\}. \] In addition, for $C$ as above and $t\in[0,1]$, set 
\[F^i(C,t)=\begin{cases}(\overline{z}_1(C,t),\ldots,\overline{z}_k(C,t)), & \mbox{if $i<k$;}\\
 C, & \mbox{if $i=k$,}
 \end{cases}\]
where $\overline{z}_j(C, t)=x_j+t(j-1)\overline{\epsilon}(C)e_C$ for $j=1,\ldots,k$. This defines a continuous ``desingularization'' deformation $F^i:A_i\times [0,1]\to F(\mathbb{R}^d,k)$ of $A_i$ into $A_k$ inside $F(\mathbb{R}^d,k)$. Note that neither the lines $L_C$ and $L^\prime_ C$ nor their orientations change under the desingularization, i.e.,~$L_{F^i(C,t)}=L_C$, $L^\prime_{F^i(C,t)}=L^\prime_ C$, and $e_{F^i(C,t)}=e_C$ for all $t\in[0,1]$.

\subsubsection{The sets $A_{ij}$ and $B_{ij}$}\label{subtn322} For $i,j=2,\ldots,k$ let 
\begin{align*}
A_{ij}:=\{(C,C^\prime)\in A_i\times A_j\colon e_C\neq -e_{C^\prime}\}\\
B_{ij}:=\{(C,C^\prime)\in A_i\times A_j\colon e_C =  -e_{C^\prime}\}
\end{align*}
The sets $A_{ij}$ and $B_{ij}$ are ENR's (for they are semi-algebraic) covering $F(\mathbb{R}^d,k)\times F(\mathbb{R}^d,k)$ that satisfy
\begin{equation}\label{clausura}
\overline{A_{ij}}\subseteq\bigcup_{r\leq i, \;s\leq j} A_{rs}\cup \bigcup_{r\leq i, \;s\leq j} B_{rs} \quad\mbox{and}\quad \overline{B_{ij}}\subseteq \bigcup_{r\leq i,\; s\leq j} B_{rs},
\end{equation} 
in view of~(\ref{cerraduras2}). We also consider subsets $X$ and $Y$ of $F(\mathbb{R}^d,k)\times F(\mathbb{R}^d,k)$ defined by
\begin{align*}
X\hspace{.6mm}&:=\{(C,C^\prime)\in F(\mathbb{R}^d,k)\times F(\mathbb{R}^d,k) \colon e_C\neq -e_{C^\prime}\mbox{ with both $C$ and $C'$ colinear}\},\\
Y\hspace{.6mm}&:=\{(C,C^\prime)\in F(\mathbb{R}^d,k)\times F(\mathbb{R}^d,k) \colon e_C=-e_{C^\prime}\mbox{ with both $C$ and $C'$ colinear}\},
\end{align*}
as well as subsets $X'\subset X$ and $Y'\subset Y$ defined by
\begin{align*}
X'&:=\{(C,C^\prime)\in X \colon L_{C}=L'_{C}=L'_{C'}=L_{C'} \mbox{ \;and\; }e_C=e_{C^\prime} \},\\
Y'&:=\{(C,C^\prime)\in Y \colon L_{C}=L'_{C}=L'_{C'}=L_{C'}\}.\\
\end{align*}
Here a configuration $C\in F(\mathbb{R}^d,k)$ is colinear if in fact $C\in F(L_C,k)$. Note that $X\cup Y$ is the set of all pairs of colinear configurations, whereas $X'\cup Y'$ is the subset of colinear configurations $(C,C')$ such that $L_C$ and $L_{C'}$ agree and pass through the origin.

\subsubsection{Deformations $\sigma_{ij}$} Next we define homotopies \[\sigma_{ij}:A_{ij}\times [0,1]\to F(\mathbb{R}^d,k)\times F(\mathbb{R}^d,k)\text{ and } \sigma_{ij}^\prime:B_{ij}\times [0,1]\to F(\mathbb{R}^d,k)\times F(\mathbb{R}^d,k),\] deforming $A_{ij}$ into $X$ and $B_{ij}$ into $Y$ respectively, i.e.,~such that \begin{enumerate}
    \item $\sigma_{ij}((C,C^\prime),0)=(C,C^\prime)$ and $\sigma_{ij}((C,C^\prime),1)\in X$,
    \item $\sigma_{ij}^\prime((C,C^\prime),0)=(C,C^\prime)$ and $\sigma_{ij}^\prime((C,C^\prime),1)\in Y$.
\end{enumerate} 

\smallskip\noindent\emph{The deformation $\sigma_{ij}$:} Given  a  pair $(C,C^\prime)\in A_{ij}$,  we  apply  first  the  desingularization  deformations $F^i(C,t)$ and $F^j(C^\prime,t)$ in order to take the pair $(C,C^\prime)$ into a pair of configurations $(C_1,C^\prime_1){}\in A_{kk}$ (recall $L_{C_1}=L_C$ and $L_{C_1^\prime}=L_{C^\prime}$). Next we apply the analogue of the linear deformation~(\ref{linear-transformation}), \red{with $p_{C_1}$ and $p_{C'_1}$ replacing $p$,} in order to take the pair $(C_1,C^\prime_1)$ into a pair of colinear configurations $(C_2,C^\prime_2){}\in X$. The deformation $\sigma_{ij}$ is the concatenation of the two deformations just described.

\smallskip\noindent\emph{The deformation $\sigma'_{ij}$:} Given  a  pair $(C,C^\prime)\in B_{ij}$,  we  apply  first  the  desingularization  deformations $F^i(C,t)$ and $F^j(C^\prime,t)$ in order to take the pair $(C,C^\prime)$ into a pair of configurations $(C_1,C^\prime_1){}\in B_{kk}$.  Next we apply the analogue of the linear deformation (\ref{linear-transformation}) in order to take the pair $(C_1,C^\prime_1)$ into a pair of colinear configurations $(C_2,C^\prime_2){}\in Y$. The deformation $\sigma_{ij}^\prime$ is the corresponding concatenated deformation.

\subsubsection{Deformations $\sigma$ and $\sigma'$} Next we deform $X$ into $X^\prime$ and $Y$ into $Y'$ by homotopies $\sigma: X\times [0,1]\to F(\mathbb{R}^d,k)\times F(\mathbb{R}^d,k)$ and $\sigma': Y\times [0,1]\to F(\mathbb{R}^d,k)\times F(\mathbb{R}^d,k)$ defined as follows. Let $(C,C')$ be a pair of colinear configurations in~$X$ (so $e_{C}\neq-e_{C'}$). First, making parallel translation, we deform $(C,C')$ into a pair of colinear configurations $(C_1,C_1')\in X$ for which $L_{C_1}=L'_{C_1}$ and $L_{C'_1}=L'_{C'_1}$, i.e., so that both lines $L_{C_1}$ and $L_{C_1^\prime}$ pass through the origin $0\in \mathbb{R}^d$ (note that $e_C=e_{C_1}$ and $e_{C'}=e_{C'_1}$). We then view $e_C$ and $e_{C^\prime}$ as points of the unit sphere $S^{d-1}\subset\mathbb{R}^d$ and, since they are not antipodal, we have the minimal-length geodesic path in $S^{d-1}$ $e:[0,1]\to S^{d-1},$ $$e(t)=\frac{(1-t)e_{C'}+te_{C}}{\parallel (1-t)e_{C'}+te_{C}\parallel},$$ joining $e_{C'}$ to $e_{C}$. This describes a rotation (pivoting at the origin) of the line $L_{C'_1}$ towards the line $L_{C_1}$ which ``drags'' $C'_1$ into a linear configuration $C_2$ with $L_{C_2}=L_{C_1}$ and $e_{C_2}=e_{C_1}$. This produces a deformation of $(C_1,C'_1)$ into the pair of colinear configurations $(C_1,C_2)\in X'$. The desired homotopy $\sigma$ is the resulting concatenated deformation.

\smallskip
The homotopy $\sigma'$ is defined analogously but in a simpler manner, as we do not need the second half of the deformation used in the case of $\sigma$. Indeed, we only need the portion of the deformation coming from parallel translation in order to define $\sigma'$.

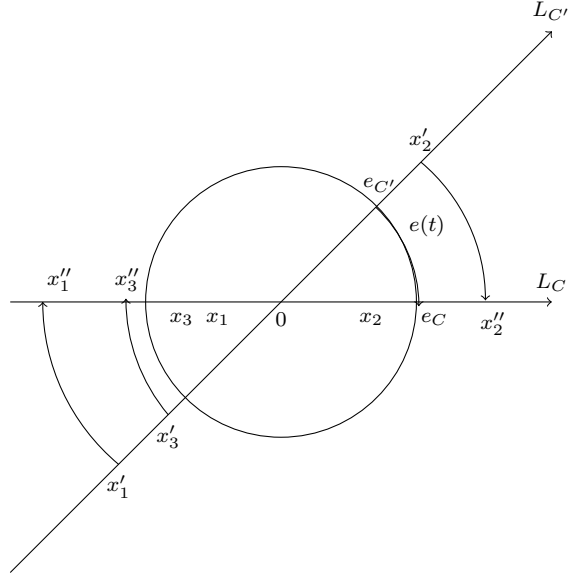
\begin{figure}[htb]
 \centering
 \begin{tikzpicture}[x=.6cm,y=.6cm]
\draw[->](-6,0)--(6,0); \draw[->](-6,-6)--(6,6); \node[above] at (6,0) {\tiny$L_C$}; \node[above] at (6,6) {\tiny$L_{C^\prime}$}; \node[below] at (0,0) {\tiny$0$};
\node[below] at (3.4,0) {\tiny$e_C$}; \node[above] at (2.2,2.2) {\tiny$e_{C^\prime}$};
\draw[black](0,0) circle (3);
\draw[black, ->] (2.1,2.1) arc (47:0:3); \node[anchor=west] at (2.6,1.7) {\tiny$e(t)$};
\draw[black, ->] (3.1,3.1) arc (50:0:4); \node[above] at (3.1,3.1) {\tiny$x_2^\prime$}; \node[below] at (4.7,0) {\tiny$x_2^{\prime\prime}$};
\draw[black, ->] (-2.5,-2.5) arc (220:180:4); \node[above] at (-3.4,0) {\tiny$x_3^{\prime\prime}$}; \node[below] at (-2.5,-2.5) {\tiny$x_3^{\prime}$};
\draw[black, ->] (-3.6,-3.6) arc (230:180:4.7); \node[above] at (-4.9,0) {\tiny$x_1^{\prime\prime}$}; \node[below] at (-3.6,-3.6) {\tiny$x_1^{\prime}$};
\node[below] at (-2.2,0) {\tiny$x_3$}; \node[below] at (-1.4,0) {\tiny$x_1$}; \node[below] at (2,0) {\tiny$x_2$};
\end{tikzpicture}
 \caption{The second portion of the deformation $\sigma$.}
 \label{rotatio}
\end{figure} 

\subsubsection{Section over $\mathcal{C}$}\label{sbsctn325}
Let $\mathcal{C}\subset F(\mathbb{R}^d,k)\times F(\mathbb{R}^d,k)$ be the set of pairs $(C,C^\prime)$ of colinear configurations such that $L_C=L_{C^\prime}=:L_{C,C'}$. Formula~(\ref{Gamma1}) defining the motion planning algorithm $\Gamma$ at the beginning of our revision of Giese-Mas' motion planning algorithm is readily adaptable to yield a continuous motion planning algorithm $$\overline{\Gamma}\colon\mathcal{C}\to P F(\mathbb{R}^d,k)$$ provided $d$ is even (this is the only place where the hypothesis about the parity of $d$ is used). Informally---but rather transparently---, the $e_1$ axis in Figure~\ref{algorithm1} is replaced by the common line $L_{C,C'}$ oriented via $e_C$, whereas the ``shifting'' direction $e_2$ in Figure~\ref{algorithm1} is replaced by $v(e_C)$. Here $v$ denotes a fixed unitary tangent vector field on $S^{d-1}$, say $v(x_1,y_1,\ldots,x_\ell,y_\ell)=(-y_1,x_1,\ldots,-y_\ell,x_\ell)$ with $d=2\ell$. Explicitly, if $C=(x_1,\ldots,x_k)$ and $C^\prime=(x^\prime_1,\ldots,x^\prime_k)$, then the path $\overline{\Gamma}(C,C')$ in $F(\mathbb{R}^d,k)$ from $C$ to $C'$ has components $(\overline{\Gamma}^{C,C'}_1,\ldots,\overline{\Gamma}^{C,C'}_k)$ defined by
    \[\overline{\Gamma}^{C,C'}_{i}(t)=\begin{cases}
  \begin{array}{ll}
    x_{i}+\red{(3ti)v(e_C),} & \hbox{for $0\leq t\leq \frac{1}{3}$;} \\
    x_{i}+iv(e_C)+(3t-1)(x^\prime_{i}-x_{i}), & \hbox{for $\frac{1}{3}\leq t\leq \frac{2}{3}$;}\\
    x^\prime_{i}+i(3-3t)v(e_C), & \hbox{for $\frac{2}{3}\leq t\leq 1$.}
    \end{array}
\end{cases}\] 
Since $X^\prime{}\cup Y' \subset\mathcal{C}$, the restriction of $\overline{\Gamma}$ yields continuous motion planning algorithms on $X'$ as well as on $Y'$.

\subsubsection{Repacking regions of continuity} 
As explained in Remark \ref{constructing-sections-via-deformations-higher-case}, we can combine the continuous motion planning algorithm $\overline{\Gamma}$ with the concatenation of the deformations discussed so far to obtain continuous motion planning algorithms
\begin{equation}\label{porhoy}
A_{i,j}\to P F(\mathbb{R}^d,k) \quad \mbox{and}\quad B_{i,j}\to P F(\mathbb{R}^d,k),
\end{equation}
for $i,j=2,\ldots,k$. The corresponding upper bound TC$(F(\mathbb{R}^d,k))\leq2(k-1)^2$ is improved by repacking these regions of continuity. Set
\[W_{\ell}=\bigcup_{i+j=\ell}A_{ij}\cup \bigcup_{r+s=\ell+1}B_{rs}\] for $\ell=3,\ldots,2k$. For instance $W_3=B_{2,2}$. In view of (\ref{clausura}), the sets assembling each $W_\ell$ are topologically disjoint, so the sets $W_\ell$ are ENR's covering $F(\mathbb{R}^d,k)\times F(\mathbb{R}^d,k)$ on each of which the corresponding algorithms in~(\ref{porhoy}) assemble a continuous motion planning algorithm. We have thus constructed a tame motion planning algorithm in $F(\mathbb{R}^d,k)$ having $2k-2$ regions of continuity $W_3,W_4,\ldots,W_{2k}$.


\section{A higher tame motion planning algorithm in $F(\mathbb{R}^d,k)$}\label{seccion4seqalgorithm}

In this section we present two optimal tame $n$-th sequential motion planning algorithms in $F(\mathbb{R}^d,k)$, which generalize in a natural way the algorithms presented in the previous section. As indicated in the introduction, the first algorithm has $n(k-1)+1$ regions of continuity, works for any $d,k,n\geq 2$, and is optimal when $d$ is odd. The second algorithm, \red{which is defined when $d$ is even}, has $n(k-1)$ regions of continuity and is optimal. The algorithms we present in this section can be used in designing practical systems controlling sequential motion of many objects moving in Euclidean space without collisions.

\subsection{A higher motion planning algorithm in $F(\mathbb{R}^d,k)$ for any $d\geq 2$}

A version of the algorithm developed in this subsection is the topic in Borat's work~\cite{borat}. As explained in Remark~\ref{muchasparticulas}, our version is more convenient for implementation purposes.

\subsubsection{Section over $F(\mathbb{R},k)^n=F(\mathbb{R},k)\times\cdots\times F(\mathbb{R},k)$}

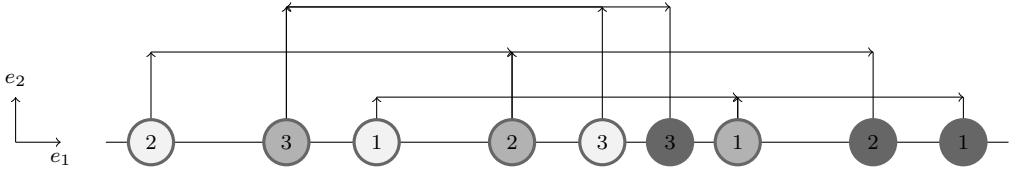
\begin{figure}[h]
 \centering
\begin{tikzpicture}[x=.6cm,y=.6cm]
\draw[->](-2,0)--(-1,0);  
\draw[->](-2,0)--(-2,1);  
\draw(0,0)--(20,0); 
\node [below] at (-1,0) {\tiny$e_1$};
\node [above] at (-2,1) {\tiny$e_2$};
 \draw[->](6,0)--(6,1);  \draw[->](6,1)--(14,1); \draw[->](14,1)--(14,0);
 \draw[->](14,0)--(14,1); \draw[->](14,1)--(19,1); \draw[->](19,1)--(19,0);
\filldraw[color=black!60, fill=black!5, very thick](6,0) circle (0.5); \node[ ] at (6,0) {\tiny$1$}; \filldraw[color=black!60, fill=black!30, very thick](14,0) circle (0.5); \node[ ] at (14,0) {\tiny$1$};
\filldraw[color=black!60, fill=black!60, very thick](19,0) circle (0.5); \node[ ] at (19,0) {\tiny$1$};
\draw[->](1,0)--(1,2); \draw[->](1,2)--(9,2); \draw[->](9,2)--(9,0);
\draw[->](9,0)--(9,2); \draw[->](9,2)--(17,2); \draw[->](17,2)--(17,0);
\filldraw[color=black!60, fill=black!5, very thick](1,0) circle (0.5); \node[ ] at (1,0) {\tiny$2$}; \filldraw[color=black!60, fill=black!30, very thick](9,0) circle (0.5); \node[ ] at (9,0) {\tiny$2$};
\filldraw[color=black!60, fill=black!60, very thick](17,0) circle (0.5); \node[ ] at (17,0) {\tiny$2$};
\draw[->](11,0)--(11,3); \draw[->](11,3)--(4,3); \draw[->](4,3)--(4,0);
\draw[->](4,0)--(4,3); \draw[->](4,3)--(12.5,3); \draw[->](12.5,3)--(12.5,0);
\filldraw[color=black!60, fill=black!5, very thick](11,0) circle (0.5); \node[ ] at (11,0) {\tiny$3$}; \filldraw[color=black!60, fill=black!30, very thick](4,0) circle (0.5); \node[ ] at (4,0) {\tiny$3$};
\filldraw[color=black!60, fill=black!60, very thick](12.5,0) circle (0.5); \node[ ] at (12.5,0) {\tiny$3$};
\end{tikzpicture}
\caption{Section over $F(\mathbb{R},k)^n$.}
 \label{algorithm1seq}
\end{figure}

Recall we take the standard embedding $\mathbb{R}:=\{(x,0,\ldots,0)\in \mathbb{R}^d:~x\in \mathbb{R}\}$, so that $F(\mathbb{R},k)$ is naturally a subspace of $F(\mathbb{R}^d,k)$. The motion planning algorithm $\Gamma:F(\mathbb{R},k)\times F(\mathbb{R},k)\to P F(\mathbb{R}^d,k)$ given by~(\ref{Gamma1}) yields a continuous $n$-th motion planning algorithm \[\Gamma_n:F(\mathbb{R},k)\times\cdots\times F(\mathbb{R},k)\to P F(\mathbb{R}^d,k)\] given by concatenation of paths (see Figure~\ref{algorithm1seq}.) \begin{equation}
    \Gamma_n(C_1,\ldots,C_n)=\Gamma(C_1,C_2)\cdot\cdots\cdot\Gamma(C_{n-1},C_{n}).
\end{equation}

\medskip





\subsubsection{Motion planning algorithms $\sigma_{j_1,\ldots,j_n}$} We now go back to the notation introduced in Subsection~\ref{subseccion312} where, for $1\leq i\leq k$, we constructed ENR's $A_i$ covering $F(\mathbb{R}^d,k)$, as well as concatenated homotopies $A_i\times [0,1]\to F(\mathbb{R}^d,k)$ deforming $A_i$ into $F(\mathbb{R},k)$. Together with the motion planning algorithm $\Gamma_n$, these deformations yield, by Remark~\ref{constructing-sections-via-deformations-higher-case}, continuous $n$-th motion planning algorithms 
\[\sigma_{j_1,\ldots,j_n}:A_{j_1}\times\cdots\times A_{j_n}\to P F(\mathbb{R}^d,k), \quad j_1,\ldots,j_n=1, 2,\ldots,k.\] Indeed, the desingularization deformation $D^{j_1}\times\cdots\times D^{j_n}$ takes $A_{j_1}\times\cdots\times A_{j_n}$ into $A_k^n$; then we apply the deformation $\varphi\times\cdots\times\varphi \;\, (n-\text{times})$ which takes $A_k^n$ into $F(\mathbb{R},k)^n$; and finally we apply Remark \ref{constructing-sections-via-deformations-higher-case}. Let us emphasise that the above description of $\sigma_{j_1,\ldots,j_n}$ is fully implementable.

\subsubsection{Combining regions of continuity.} The ENR's $A_{j_1}\times\cdots\times A_{j_n}$, $j_1,\ldots,j_n=1, 2,\ldots,k$, are mutually disjoint and cover the whole product $F(\mathbb{R}^d,k)^n$. The resulting estimate TC$_n(F(\mathbb{R}^d,k))\leq k^n$ coming from Proposition~\ref{rudi} and the motion planning algorithms $\sigma_{j_1,\ldots,j_n}$ \red{are now} improved by combining the domains of continuity to yield $n(k-1)+1$ covering ENR's $W_{\ell}$, $\ell=n,n+1,\ldots,nk$, each admitting a continuous $n$-th motion planning algorithm. Explicitly, let \begin{equation}
    W_{\ell}=\bigcup_{j_1+\cdots+j_n=\ell}A_{j_1}\times\cdots\times A_{j_n},
\end{equation}
where $\ell=n,n+1,\ldots,nk$. By~(\ref{cerraduras1}), any two distinct $n$-tuples $(j_1,\ldots,j_n)$  and $(j^\prime_1,\ldots,j^\prime_n)$  with $j_1+\cdots+j_n=j^\prime_1+\cdots+j^\prime_n$ determine topologically disjoint sets $A_{j_1}\times\cdots\times A_{j_n}$ and $A_{j^\prime_1}\times\cdots\times A_{j^\prime_n}$ in $F(\mathbb{R}^d,k)^n$, i.e., $\overline{A_{j_1}\times\cdots\times A_{j_n}}\cap (A_{j^\prime_1}\times\cdots\times A_{j^\prime_n})=\varnothing$. Therefore the motion planning algorithms $\sigma_{j_1,\ldots,j_n}$ with $j_1+\cdots +j_n=\ell\hspace{.3mm}$ jointly  define  a  continuous motion planning algorithm on $W_{\ell}$. We have thus constructed a tame $n$-th sequential motion planning algorithm in $F(\mathbb{R}^d,k)$ having $n(k-1)+1$ domains of continuity $W_n,W_{n+1},\ldots, W_{nk}$.

\subsection{An optimal higher motion planning algorithm in $F(\mathbb{R}^d,k)$ for $d$ even}

In this section we improve the $n$-th sequential motion planning algorithm in $F(\mathbb{R}^d,k)$ of the previous section  under the assumption (in force throughout the section) that $d$ is even. The improved $n$-th motion planning algorithm has $n(k-1)$ domains of continuity, and is therefore optimal (Theorem~\ref{gongra}). This gives the higher-TC analogue of the construction in Subsection~\ref{section2}.

\subsubsection{The sets $A_{i_1,\ldots,i_n;j}$} For a configuration $C\in F(\mathbb{R}^d,k)$, we  now bring the notation $L_C$, $L'_C$, $e_C$, $p_C$ and $\overline{\text{cp}}(C)$ in Subsection~\ref{section2} back to use. For $i_1,\ldots,i_n{}\in\{2,\ldots,k\}$ and $j{}\in\{0,1,\ldots,n-1\}$ we denote by $A_{i_1,\ldots,i_n;j}$ the set of all $n$-tuples of configurations $(C_1,\ldots,C_n)\in F(\mathbb{R}^d,k)^n$ satisfying
\begin{itemize}
\item $\overline{\text{cp}}(C_s)=i_s$ for $s=1,\ldots,n$, and
\item the $n$-tuple $(e_{C_1},\ldots,e_{C_n})$ has exactly $j$ antipodes to $e_{C_1}$.
\end{itemize}
The sets $A_{i_1,\ldots,i_n;j}$ are pairwise disjoint ENR's covering $F(\mathbb{R}^d,k)^n$. As in Subsection~\ref{section2}, the goal is to construct a continuous $n$-th motion planning algorithm on each $A_{i_1,\ldots,i_n;j}$, and then make a suitable repacking of these domains.

\begin{example} For $n=2$, we have $A_{i,j;0}=A_{i,j}$ and $A_{i,j;1}=B_{i,j}$ (see Subsection~\ref{subtn322}).
\end{example}

In view of~(\ref{cerraduras2}), for $i_1,\ldots,i_n$ and $j$ as above, we have
\begin{equation}\label{clausura2}
\overline{A_{i_1,\ldots,i_n;j}} \subset \bigcup_{r_1\leq i_1,\ldots, \hspace{.4mm}r_n\leq i_n, \hspace{.4mm}s\geq j} A_{r_1,\ldots,r_n;s}.
\end{equation} 

\subsubsection*{The sets $X_j$ and $X^\prime_j$} For $0\leq j\leq n-1$, let $X_j\subset F(\mathbb{R}^d,k)^n$ denote the set of all $n$-tuples of colinear configurations $(C_1,\ldots,C_n)\in F(\mathbb{R}^d,k)^n$ such that the $n$-tuple $(e_{C_1},\ldots,e_{C_n})$ has exactly $j$ antipodes to $e_{C_1}$. Consider in addition the subsets $X^\prime_j\subset F(\mathbb{R}^d,k)^n$ consisting of all $n$-tuples of colinear configurations $(C_1,\ldots,C_n)\in X_j$ such that $L_{C_i}=L'_{C_i}$ and $L_{C_i}=L_{C_1}$ for all $i\in\{1,\ldots,n\}$. 

\subsubsection{Deformations $\sigma_{i_1,\ldots,i_n;j}$} Next we define homotopies
\[\sigma_{i_1,\ldots,i_n;j}:A_{i_1,\ldots,i_n;j}\times [0,1]\to F(\mathbb{R}^d,k)^n\] deforming $A_{i_1,\ldots,i_n;j}$ into $X_j$, i.e.,~such that \[\sigma_{i_1,\ldots,i_n;j}((C_1,\ldots,C_n),0)=(C_1,\ldots,C_n) \text{ \ \ and \ \ } \sigma_{i_1,\ldots,i_n;j}((C_1,\ldots,C_n),1)\in X_j.\] 
Explicitly, given an $n$-tuple $(C_1,\ldots,C_n)\in A_{i_1,\ldots,i_n;j}$,  we  apply  first  the $n$-tuple of desingularization  deformations $(F^{i_1}(C_1,t), F^{i_2}(C_2,t), \cdots, F^{i_n}(C_{n},t))$ in order to take the $n$-tuple $(C_1,\ldots,C_n)$ into an $n$-tuple of configurations $(C_1^\prime,\ldots,C_n^\prime)\in A_{k,\ldots,k;j}$ (note this yields $L_{C'_i}=L_{C_i}$ with $e_{C'_i}=e_{C_i}$). Next we apply the corresponding analogues of the linear deformation (\ref{linear-transformation}) in order to take the $n$-tuple $(C_1^\prime,\ldots,C_n^\prime)$ into an $n$-tuple of colinear configurations $(C_1^{\prime\prime},\ldots,C_n^{\prime\prime})\in X_j$ (once again $L_{C''_i}=L_{C_i}$ with $e_{C''_i}=e_{C_i}$). The deformation $\sigma_{i_1,\ldots,i_n;j}$ is the concatenation of the two deformations just described.

\subsubsection{Deformation $\sigma_j$}
Homotopies $\sigma_j\colon X_j\times [0,1]\to F(\mathbb{R}^d,k)^n$, for $0\leq j\leq n-1$, deforming $X_j$ into $X_j^\prime$ are defined next. Let $(C_1,\ldots,C_n)$ be an $n$-tuple of colinear configurations in~$X_j$. First, making parallel translation, we deform $(C_1,\ldots,C_n)$ into an $n$-tuple of colinear configurations $(C'_1,\ldots,C'_n)\in X_j$ for which each line $L_{C'_i}$ passes through the origin $0\in \mathbb{R}^d$ (note that this is done so that $e_{C_i}=e_{C'_i}$ for $1\leq i\leq n$). Continuity on $(C_1,\ldots, C_n)$ of this deformation is obvious. We then view each $e_{C_i}$ as a point of the unit sphere $S^{d-1}\subset\mathbb{R}^d$ and, whenever $e_{C_i}$ is not antipodal to $e_{C_1}$, we have the minimal-length geodesic path in $S^{d-1}$, $e_i:[0,1]\to S^{d-1},$ $$e_i(t)=\frac{(1-t)e_{C_i}+te_{C_1}}{\parallel (1-t)e_{C_i}+te_{C_1}\parallel},$$ joining $e_{C_i}$ to $e_{C_1}$. This describes a rotation (pivoting at the origin) of the line $L_{C'_i}$ towards the line $L_{C'_1}$ which ``drags'' $C'_i$ into a linear configuration $C''_i$ with $L_{C''_i}=L_{C'_1}$ and $e_{C''_i}=e_{C'_1}$. This produces a deformation of $(C'_1,\ldots,C'_n)$ into an $n$-th tuple of colinear configurations $(C''_1,\ldots,C''_n)\in X'_j$, where we set $C''_i=C'_i$ whenever $e_{C_i}$ and $e_{C_1}$ are antipodal, in which case the ``deformation'' of $C'_i$ into $C''_i$ is stationary. Continuity on $(C'_1,\ldots,C'_n)$ of this second deformation holds because it does not leave the (fixed) domain $X_j$. The desired homotopy $\sigma_j$ is the resulting concatenated deformation.

\subsubsection{An $n$-th motion planning algorithm on each $A_{i_1,\ldots,i_n;j}$} 

In Subsection~\ref{sbsctn325} we constructed a continuous motion planning algorithm $\overline{\Gamma}:\mathcal{C}\to P F(\mathbb{R}^d,k)$ on the set $\mathcal{C}\subset F(\mathbb{R}^d,k)\times F(\mathbb{R}^d,k)$ consisting of all pairs $(C,C^\prime)$ of colinear configurations such that $L_C=L_{C^\prime}$. More generally, we now let 
$\mathcal{C}(n)\subset F(\mathbb{R}^d,k)^n$ stand for the set of all $n$-tuples $(C_1,\ldots,C_n)$ of colinear configurations such that $L_{C_1}=\cdots=L_{C_n}=:L_{C_1,\ldots,C_n}$. Then a continuous $n$-th motion planning algorithm $\overline{\Gamma}_n:\mathcal{C}(n)\to P F(\mathbb{R}^d,k)$ is given by concatenation of paths,\begin{equation*}
    \Gamma_n(C_1,\ldots,C_n)=\Gamma(C_1,C_2)\cdot\cdots\cdot\Gamma(C_{n-1},C_{n}).
\end{equation*}
Since each $X'_j$ is a subset of $\mathcal{C}(n)$, the deformations discussed so far yield, in view of Remark~\ref{constructing-sections-via-deformations-higher-case}, $n$-th continuous motion planning algorithms $\sigma_{i_1,\ldots,i_n;j}:A_{i_1,\ldots,i_n;j}\to P F(\mathbb{R}^d,k)$ for $i_1,\ldots,i_n\in\{2,\ldots,k\}$ and $j\in\{0,1,\ldots,n-1\}$. 

\subsubsection{Repacking regions of continuity} The sets $A_{i_1,\ldots,i_n;j}$ are pairwise disjoint ENR's covering $F(\mathbb{R}^d,k)^n$ on each of which we have constructed a continuous $n$-th motion planning algorithm. The upper bound TC$_n(F(\mathbb{R}^d,k))\leq n(k-1)^n$ given by Proposition~\ref{rudi} is next improved with a suitable repacking of the domains $A_{i_1,\ldots,i_n;j}$. Set 
\[W_{\ell}=\bigcup_{r_1+\cdots+r_n=\ell}\hspace{-2mm}A_{r_1,\ldots,r_n;0}\hspace{2.5mm}\cup \bigcup_{r_1+\cdots+r_n=\ell+1}\hspace{-2.5mm}A_{r_1,\ldots,r_n;1}\hspace{2.5mm}\cup\cdots\cup \bigcup_{r_1+\cdots+r_n=\ell+n-1}\hspace{-4.5mm}A_{r_1,\ldots,r_n;n-1}\] for $\ell=n+1, \ldots, 2n-1,2n,\ldots,nk$. For instance,
\begin{eqnarray*}
W_{n+1} = A_{2,\ldots,2;n-1} \quad \mbox{and}\quad
W_{n+2} = A_{2,\ldots,2;n-2}\;\;\;\cup \bigcup_{r_1+\cdots+r_n=2n+1}\hspace{-3mm}A_{r_1,\ldots,r_n;n-1}.
\end{eqnarray*}
From (\ref{clausura2}), the sets assembling each $W_{\ell}$ are topologically disjoint, so the various sets $W_{\ell}$ are pairwise disjoint ENR's covering $F(\mathbb{R}^d,k)^n$ on each of which the corresponding algorithms $\sigma_{i_1,\ldots,i_n;j}$ assemble a continuous $n$-th motion planning algorithm. We have thus constructed a (global) tame $n$-th sequential motion planning algorithm in $F(\mathbb{R}^d,k)$ having $n(k-1)$ regions of continuity $W_{n+1},W_{n+2},\ldots, W_{nk}$.

\bibliographystyle{plain}

\end{document}